\documentclass[runningheads]{llncs}
\usepackage{graphicx}

\usepackage{tikz}
\usepackage{comment}
\usepackage{amsmath,amssymb} %
\usepackage{color}
\usepackage{booktabs}
\usepackage{tabularx}
\usepackage{cuted}
\usepackage{multirow}
 \usepackage[pagebackref,breaklinks,colorlinks]{hyperref}
\usepackage[numbers,sort,compress]{natbib}
\usepackage{wrapfig,lipsum}

\usepackage[accsupp]{axessibility}  %

\usepackage{subcaption}

\usepackage{overpic}
\usepackage{enumitem} %
\usepackage{overpic} %
\usepackage{color}

\definecolor{turquoise}{cmyk}{0.65,0,0.1,0.3}
\definecolor{purple}{rgb}{0.65,0,0.65}
\definecolor{dark_green}{rgb}{0, 0.5, 0}
\definecolor{orange}{rgb}{0.8, 0.6, 0.2}
\definecolor{red}{rgb}{0.8, 0.2, 0.2}
\definecolor{darkred}{rgb}{0.6, 0.1, 0.05}
\definecolor{blueish}{rgb}{0.0, 0.3, .6}
\definecolor{light_gray}{rgb}{0.7, 0.7, .7}
\definecolor{pink}{rgb}{1, 0, 1}
\definecolor{greyblue}{rgb}{0.25, 0.25, 1}

\usepackage{blindtext}

\renewcommand{\paragraph}[1]{\vspace{1em}\noindent\textbf{#1}.}

\newcommand{\shapecoeff}{\boldsymbol{\beta}}
\newcommand{\shapedim}{{\left| \shapecoeff \right|}}

\newcommand{\posecoeff}{\boldsymbol{\theta}}

\newcommand{\expcoeff}{\boldsymbol{\psi}}
\newcommand{\expdim}{{\left| \expcoeff \right|}}

\newcommand{\numjoints}{k}

\newcommand{\lighting}{\mathbf{\gamma}}

\newcommand{\albedo}{A}
\newcommand{\albedocoeffs}{\boldsymbol{\alpha}}
\newcommand{\albedodim}{\left| \albedocoeffs \right|}

\newcommand{\uvsize}{d}

\newcommand{\normals}{N}

\newcommand{\modelname}{TRUST\xspace} 
\newcommand{\datasetname}{FAIR\xspace}

\newcommand{\eg}{e.g.~} 
\newcommand{\etal}{et al.\xspace}

\begin{document}

\pagestyle{headings}
\mainmatter

\title{Towards Racially Unbiased Skin Tone Estimation via Scene Disambiguation}

\authorrunning{H. Feng et al.}

\author{
Haiwen Feng \and
Timo Bolkart \and
Joachim Tesch \and
Michael J. Black \and\\
Victoria Abrevaya
}

\institute{Max Planck Institute for Intelligent Systems, T{\"u}bingen, Germany\\
\email{\{hfeng,tbolkart,jtesch,black,vabrevaya\}@tuebingen.mpg.de}}

\maketitle

\vspace{-0.35in}
\begin{figure}
    \centerline{
     \includegraphics[width=1\textwidth]{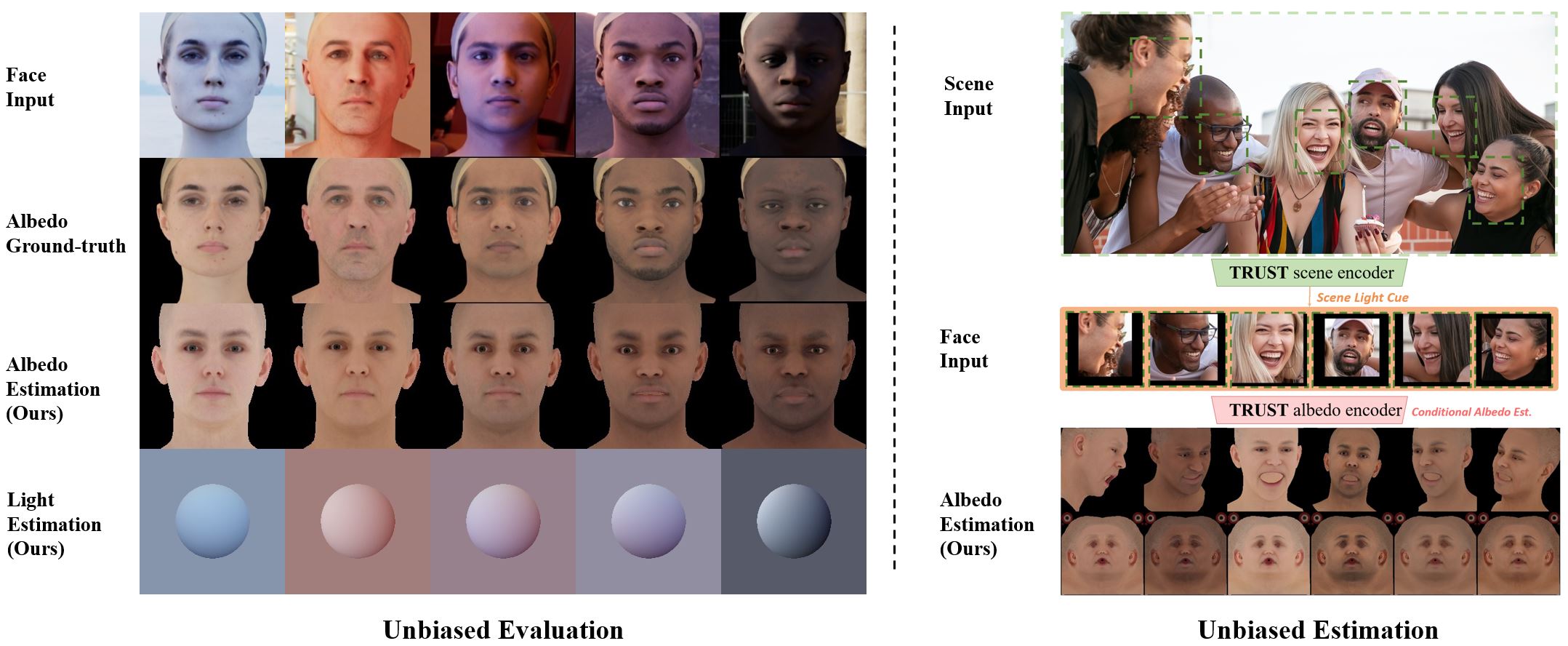}}
    \caption{{\bf \datasetname} (left) is a new  dataset for unbiased albedo estimation that uses high-quality facial scans, varied lighting, and a new evaluation metric to benchmark current approaches in terms of accuracy and fairness. We also propose a new network, {\bf \modelname} (right), for facial albedo estimation that reduces bias by addressing the light/albedo ambiguity  using scene disambiguation cues.}
    \label{fig:teaser}
\end{figure}\
\vspace{-0.5in}

\begin{abstract}

Virtual facial avatars will play an increasingly important role in immersive communication, games and the metaverse, and it is therefore critical that they be inclusive. 
This requires accurate recovery of the albedo, 
regardless of age, sex, or ethnicity. 
While significant progress has been made on %
estimating 3D facial geometry, 
appearance estimation has received less attention.
The task is fundamentally ambiguous because the observed color is a function of albedo and lighting, both of which are unknown.
We find that current methods are biased towards light skin tones due to (1) strongly biased priors that prefer lighter pigmentation and 
(2) algorithmic solutions that disregard the light/albedo ambiguity.
To address this, we propose a new evaluation dataset (\datasetname) and an algorithm (\modelname) to improve albedo estimation and, hence, fairness. 
Specifically, we create the first facial albedo evaluation benchmark where subjects are balanced in terms of skin color, and measure accuracy using the Individual Typology Angle (ITA) metric. 
We then address  the light/albedo ambiguity by building on a key observation:
the image of the full scene 
--as opposed to a cropped image of the face-- contains important information about lighting that can be used for disambiguation. 
\modelname regresses facial albedo by conditioning on both the face region and a global illumination signal obtained from the scene image. 
Our experimental results show significant improvement compared to state-of-the-art methods on albedo estimation, both in terms of accuracy and fairness. 
The evaluation benchmark and code are available
for research purposes at \href{https://trust.is.tue.mpg.de}{https://trust.is.tue.mpg.de}.

\end{abstract}
\section{Introduction}
\label{sec:intro}

For critical systems such as face recognition, automated decision making or medical diagnosis, the development of machine-learning-based methods has been followed by questions about how to make their decisions fair to all sectors of the population \cite{Adamson18,BuolamwiniGebru18,Kinyanjui20,Osoba17}. 
For example, Buolamwini and Gebru~\cite{BuolamwiniGebru18} identify race and gender biases in face analysis methods, which disproportionately misclassify dark-skinned females; Rajkomar et al.~\cite{Rajkomar18} note an influence of historical biases in machine learning methods for health care; and Kim et al.~\cite{Kim21} point out how rendering algorithms in computer graphics are designed mostly for light-skinned people.

However, no such analysis exists for the task of 3D facial avatar creation.
Here the problem involves estimating the 3D geometry and albedo from one or more images, where albedo is defined as a texture map corresponding to the diffuse reflection of the face (including skin tone, lips, eye, etc). With growing interest in on-line virtual communication, gaming, and the metaverse, the role of %
facial avatars in our lives will likely increase.
It is then critical that such technology is equally accessible to all and that every person can be represented faithfully, independent of  gender, age, or skin color. 

In this paper
we address, for the first time, the problem of fairness in 
3D avatar creation from images in-the-wild, 
and show that current methods are %
biased towards 
estimating albedo with a light skin color. 
While skin-tone bias has been extensively studied in the field of face recognition 
\cite{Drozdowski20, Terhorst21}, here
we examine a different problem.
Specifically, we consider bias within methods that \emph{regress} facial albedo (of which skin tone is only one aspect) and discuss the particular challenges that this involves. 
We analyze three main sources for this bias. 
First, existing statistical albedo models are trained from unbalanced datasets, producing strongly biased priors.
Second, existing albedo regression methods are unable to factor lighting from albedo. 
When combined with a biased prior, they simply infer dim illumination to compensate for dark skin.
Finally, there is, to date, no standard evaluation protocol that quantitatively reveals bias in albedo estimation. 
This has led to the field largely ignoring issues of fairness in albedo estimation.

This work makes several key contributions to advance fairness in facial albedo estimation.
First, we create a new dataset of realistic synthetic images of faces in varying lighting conditions. We use this dataset, 
together with %
specifically designed
metrics, %
as a benchmark for evaluating accuracy and fairness of facial albedo reconstruction (Fig.~\ref{fig:teaser}, left); we call the benchmark \datasetname for Facial Albedo Independent of Race. 
With this we are able, for the first time, to quantify the bias of existing methods.
We find that all existing methods are biased to light skin tones, and analyze the technical reasons behind this. %

Finally, we introduce a new neural network model called \modelname (Towards Racially Unbiased Skin Tone estimation) that produces state-of-the-art results in terms of accuracy, as well as less biased albedo estimates (Fig.~\ref{fig:teaser}, right). 
The key to \modelname is the design of a novel architecture and losses that explicitly address the light/albedo ambiguity. 
Specifically, we propose a two-branch network to leverage scene cues for disambiguation. 
The first branch recovers the illumination signal in the form of Spherical Harmonics (SH) \cite{Ramamoorthi01}, both from the entire scene image and the individual facial crops. The second branch estimates diffuse albedo from the individual crops, conditioned on the intensity of the SH vector. This provides cues about the global illumination and helps the albedo network to better decide on the overall skin tone. 
The network is further coupled with a new statistical albedo model trained using a balanced sample of subjects. The proposed approach achieves $56\%$ improvement over the best alternative method when evaluated on the new \datasetname benchmark, highlighting the importance of tackling fairness both from the dataset and the algorithmic point of view.

In summary, our contributions are: 
(1) We identify, analyze and quantify the problem of biased facial albedo estimation.
(2) We propose a new synthetic benchmark, as well as new metrics that measure performance in terms of skin tone and diversity. 
(3) We propose a solution for albedo estimation that significantly improves fairness by explicitly addressing the light/albedo ambiguity problem through scene disambiguation cues. 
(4) The benchmark, code, and model are publicly available for research purposes.

\section{Related work}
\label{sec:related}

\textbf{3D face and appearance estimation.}
Face albedo estimation is typically approached as an inverse rendering problem, in which face geometry, reflectance and illumination parameters are estimated from a single image. 
Methods can be roughly categorized into optimization-based \cite{Aldrian12,Blanz99,Blanz02,Schonborn17,Thies2016} and learning-based  \cite{Chen19,Deng19,Feng21,Genova18,Kim18,Shang20,Tewari17}; see Egger \etal~\cite{Egger2020} for a review.
The majority of these estimate appearance parameters of a 3D morphable model (3DMM) \cite{Blanz99,Gerig2018_BFM17,Paysan09} trained from two-hundred white European subjects. %
This lack of diversity results in a strong bias towards light-skinned appearance when jointly estimating reflectance and lighting.

To obtain more varied appearance, 
Smith \etal~\cite{Smith20} build an albedo model from light-stage data, and Gecer \etal~\cite{Gecer19} build a neural generative model from $10K$ texture maps.  
Several other methods learn flexible appearance models directly from images, while jointly learning to reconstruct 3D shape and appearance \cite{Chaudhuri20,Lin20,Sahasrabudhe2019,Sengupta18,Shu17,Tewari18,Tewari19,Tran18,Tran19,Wen21}.
Another line of work synthesizes high-frequency appearance details from images \cite{Saito17,Yamaguchi18,Lattas20}.
All of these methods treat the face in isolation by only analyzing a tightly cropped face region.
In contrast, our approach exploits cues from the entire scene that help decouple albedo from lighting in unconstrained in-the-wild images. 

A related line of work directly estimates \emph{skin tone} from images -- i.e.~a single vector value (e.g.~RGB) that summarizes the color of the face. This can be used as a proxy for color correction \cite{Choi17,Marguier07,Bianco14} and for cosmetology purposes \cite{Kips20,Kips20b}. Here we focus on estimating the full albedo map, but evaluate the accuracy of both albedo and skin tone, with the goal of identifying potential fairness issues.

\noindent
\textbf{Disambiguating appearance and lighting.}
Given the RGB values of a pixel, the proportion of color explained by light versus  intrinsic appearance cannot be recovered without further knowledge \cite{Ramamoorthi01}. 
While the use of a statistical appearance prior limits the variability of the intrinsic face color, it does not fully resolve the ambiguity. %
To address this, Aldrian \etal~\cite{Aldrian12} regularize the light by imposing a ``gray world'' constraint that encourages the environment to be, on average, gray.
Hu \etal~\cite{Hu13} regularize the albedo by imposing symmetry, thus preventing illumination variation from strong point lights being explained by the appearance. 
Egger \etal~\cite{Egger18} learn a statistical prior of in-the-wild SH coefficients and use this to constrain the light in an inverse rendering application.
These methods impose priors or regularizers over the estimated light or appearance, using heuristics to complement the statistical face model. 
Instead of using the tightly cropped face in isolation, we estimate the environment light from the scene and use this in learning to disambiguate light and albedo.

\noindent
\textbf{Racial face bias.} 
Several works identify demographic bias (sex, age, race) for algorithms in the context of biometrics (e.g., face verification, age estimation, race classification, emotion classification, etc.); see Drozdowski \etal~\cite{Drozdowski20} for a review. 
Dedicated benchmark datasets were proposed in \cite{Wang19, Robinson20} to support the study of racial biases in face recognition. 
Wang \etal~\cite{Wang19} show bias in four commercial face recognition systems, and propose an algorithmic solution to mitigate this. 
Buolamwini and Gebru~\cite{BuolamwiniGebru18} report bias in gender classification systems correlated to skin tone and sex, where darker-skinned females are misclassified most frequently.   
Kim \etal~\cite{Kim21} describe racial bias in existing quantitative metrics used to group skin colors and hair types, and propose an evaluation scheme based on complex numbers to alleviate this bias. No other previous work has quantitatively assessed bias in facial albedo estimation methods, nor is there any existing dataset for this task. Hence, \datasetname provides an important new resource to the community.

\begin{figure*}
\centerline{ 
\includegraphics[width=\textwidth, trim={0 0 0 0}, clip=True]{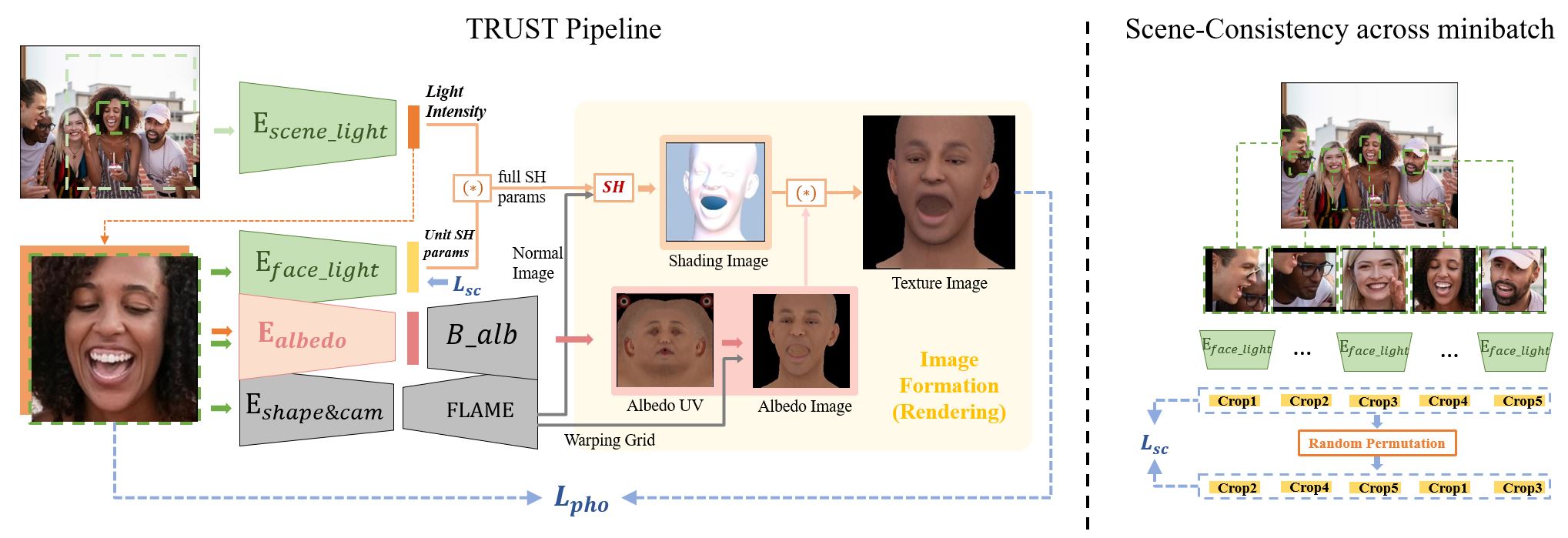}}
\vspace{-0,1in}
\caption{
We address biased albedo estimates by tackling the light/albedo ambiguity using scene disambiguation cues. Given an image of a scene with faces, we first obtain the 3-channel global light intensity factor using the $E_{scene\_light}$ encoder. We next extract facial crops, and condition the albedo encoder $E_{albedo}$ by concatenating the intensity as an extra channel to the input. Finally, a crop-based light encoder $E_{face\_light}$ estimates normalized SH parameters, which are combined with the color intensity  to obtain the final light estimate for the crop. 
The network is trained using a scene consistency loss, which requires the light from all faces in an image to be the same (i.e. permutation invariant). 
}
\label{fig:overview}
\end{figure*} %

\section{Dataset and Metrics for Quantifying Skin Tone Bias}
\label{sec:benchmark}

To develop unbiased algorithms it is important to first identify 
potential problems. 
Since it is difficult to acquire ground-truth appearance, there is to date no albedo evaluation dataset available, much less one that covers a wide range of ethnicities and scenes. 
To address this, we describe a new dataset for evaluating single-image albedo reconstruction methods, constructed using %
high-quality facial scans. 
We additionally propose a set of metrics that measure fidelity of the estimated albedo, as well as accuracy and diversity of the overall skin tone. 

\paragraph{Balancing skin type} 
There are many types of bias but here we focus on skin color, which means we need an evaluation dataset that is balanced across this factor. 
There are several ways to quantify the skin color of a person. 
While self-reported ethnicity labels are commonly used in the literature (\eg \cite{Wang19,Li20}), ethnicity is not well defined and there can be a large variety of skin types within ethnic groups.  
In dermatology, an extensively used system is the Fitzpatrick scale \cite{Fitzpatrick88}, employed also in computer vision research for balancing datasets \cite{BuolamwiniGebru18,Dooley21,Krishnapriya20,Wilson19}. 
However, the scale is based on subjective assessments of skin sensitivity and tanning ability, and it has been shown to work sub-optimally for certain populations \cite{Pichon10,Youn97}.
Instead, we employ the Individual Typology Angle (ITA) \cite{Chardon91}, 
also recently considered in~\cite{Merler19}. 
The metric is based on the L* (lightness) and B* (yellow/blue) components of the CIE L*a*b* color space, and is computed as
\begin{equation}
\label{eq:ita}
    \mathrm{ITA}(L^*, b^*) = \frac{\arctan(\frac{L^*-50}{b^*}) \times 180}{\pi}.
\end{equation}

We consider the ITA score of an albedo map to be the average of all pixel-wise ITA values within a pre-computed skin region area (defined on the UV map). ITA can be used to classify the skin according to six categories, ranging from very light (category I)  to dark (category VI) \cite{DelBino13,Chardon91}. %
It has the advantage of being an objective metric, %
easily computable from images, and significantly correlated with skin pigmentation \cite{DelBino06}. More details can be found in Sup.~Mat.

\paragraph{Dataset construction} The dataset is constructed using 206 high-quality 3D head scans purchased from Triplegangers\footnote{\url{https://triplegangers.com/}}. The scans were selected such that they cover a relatively balanced range of sexes and skin colors (ITA skin group). 
Ages range from 18 to 78 years old.
All of the scans were captured under neutral expression and uniform lighting. 
We obtain UV texture maps compatible with FLAME \cite{Li17} by registering the FLAME model to the scans 
using the approach in \cite{Li17} and we treat the texture maps as approximate ground-truth albedo.

To create a photo-realistic dataset we rendered the scans in complex scenes.
Specifically, we used 50 HDR environment maps from Poly Haven\footnote{\url{https://polyhaven.com/}} covering both natural and artificial lighting. 
In each scene we rendered three head scans under the same illumination. To ensure a balanced distribution of skin types, each image was constructed by first randomly selecting a skin type, and then randomly selecting a sample scan within the type.
Our final testing dataset contains 721 images and 2163 facial crops under different illumination, with approximate ground-truth UV maps. 
Details and examples are provided in Sup.~Mat. 
  
\paragraph{Evaluation metrics} 
To focus on both accuracy and fairness, we use the following metrics: 

    \smallskip
    \noindent \textbf{- ITA error}. We compute fidelity of skin tone by taking the average error in ITA (degrees, see Eq.~\ref{eq:ita}) between predicted and ground-truth UV maps, over a skin mask region (see Sup.~Mat.). We report average error per skin type (I to VI), as well as average ITA error across all groups. 
    
    \noindent \textbf{- Bias score}. We quantify the \emph{bias} of a method in terms of skin color by measuring the standard deviation across the \emph{per-group} ITA errors (note that this is not the same as the standard deviation over the full dataset).
    A low value indicates roughly equal performance (i.e.~unbiased) across all skin tones.
    
    \noindent \textbf{- Total score}. We summarize average ITA and bias score into a single score, which is the sum of the two. 
    
    \noindent \textbf{- MAE}. We also report results with a commonly employed image metric, namely mean average error. Here we calculate errors over the entire UV map, as opposed to only the skin region.

\paragraph{Fairness of albedo estimates of current methods} Using the above evaluation criteria, we benchmark several recent methods for facial albedo estimation from images in the wild.
These are summarized in the top rows of Tab.~\ref{tab:sota}. 
When observing the
``ITA per skin type" column, we note that the accuracy of all  methods  varies with skin type, performing  best on those values that are better represented by the statistical model\footnote{For GANFIT \cite{Gecer19}, the albedos contain a significant amount of baked-in lighting, and were captured with lower light conditions, hence the tendency to do well on dark skin tones.}. 
In particular, extreme skin types such as I and VI tend to have noticeably larger errors. 
Low (bold) numbers for some types and high ones for others on each line indicates bias. 
An unbiased algorithm should have roughly equal errors %
across skin types. %

\section{Reasons Behind Racially Biased Albedo Estimations}
\label{sec:reasons}

Recovering the 3D shape %
and appearance of an object from a single RGB image is a %
highly ill-posed problem due to fundamental ambiguities arising from the interplay between geometry \cite{Bas19,Egger21} and lighting \cite{Ramamoorthi01} in the image formation process. 
For known objects such as the human face, a standard strategy is to use a strong prior in the form of a low-dimensional statistical model, \eg the 3D Morphable Model (3DMM) \cite{Blanz99,Paysan09} and its variants \cite{Dai19,Li17,Smith20}, to constrain the solution to the space of valid shapes and appearances.
This idea has been widely adopted and has led to impressive advances in the field \cite{Egger2020}. Yet, no careful consideration has been given to ensure that the models cover a balanced demographic distribution; indeed, the most widely used appearance model \cite{Paysan09} was built using two-hundred white subjects. %
Several 3DMM variants have been made available (e.g. \cite{Smith20,Li20}) but none of them ensure a balanced distribution of skin tones.
It is worth noting that a biased albedo model, employed within a neural network, can still be trained to produce outputs that are far away from the statistical mean. This is the case for example of MGCNet \cite{Shang20} (see Tab.~\ref{tab:sota}), which uses low regularization weights %
to extrapolate results for type V skin tones. 
However, these are noisy estimates that do not faithfully represent the albedo, and the model still cannot extrapolate to type VI skin tones.

\smallskip
Even if one had an unbiased statistical model of face albedo, the problem remains ill-posed, leading to an \emph{algorithmic} source of bias \cite{Mehrabi21}.
There is a fundamental ambiguity between scene lighting and albedo that
cannot be resolved without strong assumptions  \cite{Ramamoorthi01}.
Even with a good statistical model of face albedo,
there are an infinite number of valid combinations of light and albedo that can explain the image.
For example, an image of a darked-skin person can be explained by both dark skin and a bright light, or light skin and a dim light. 
This can be easily observed by looking at the %
shading equation 
commonly used for diffuse objects: 
\begin{equation}
\label{eq:rendering}
    I_R = I_A \odot I_S,
\end{equation}

\noindent
where $I_R$ is the final rendered image, $I_S$ is the shading image, $I_A$ is the albedo image, and $\odot$ denotes the Hadamard product. %
When both the albedo and light are unknown %
there is a scale ambiguity:
for a fixed target image $I_R$, an increase in $I_S$ by a factor of $s$ results in a decrease in $I_A$ by a factor of $1/s$. Given that skin tone dominates the albedo color in $I_A$, an overly bright estimate of the light will result in an overly dark estimate of the skin tone, and vice-versa. 
To address this we next propose a method that exploits scene lighting to reduce ambiguity in conjunction with an improved prior.

\begin{table*}[t]
\centering
    \caption{Comparison to state of the art on the \datasetname benchmark. We show: average ITA error over all skin types; bias score (standard deviation); total score (avg. ITA+Bias); mean average error; and avg. ITA score per skin type in degrees (I: very light, VI: very dark). Our method achieves more balanced estimates, as can be seen in the bias score, as well as accurate skin color predictions. }
    \label{tab:sota}
    \resizebox{\linewidth}{!}{%
    \begin{tabular}{ l | c | c | c | c | cccccc }
    \toprule
    \multirow{2}{*}{Method} & \multirow{2}{*}{Avg. ITA $\downarrow$} & \multirow{2}{*}{Bias $\downarrow$} & \multirow{2}{*}{Score $\downarrow$} & \multirow{2}{*}{MAE $\downarrow$} & \multicolumn{6}{c}{ITA per skin type $\downarrow$} \\
    \cline{6-11}
    & & & & & I & II & III & IV & V & VI \\
    \midrule
    Deng \etal \cite{Deng19} & $22.57$ & $24.44$ & $47.02$   & $27.98$ & $\mathbf{8.92}$   & $9.08$   & $8.15$    & $10.90$   & $28.48$   & $69.90$ \\
    GANFIT \cite{Gecer19}   & $62.29$ & $31.81$ & $94.11$   & $63.31$ & $94.80$   & $87.83$   & $76.25$   & $65.05$   & $38.24$   & $\mathbf{11.59}$     \\
    MGCNet \cite{Shang20}  & $21.41$ & $17.58$ & $38.99$   & $25.17$  & $19.98$   & $12.76$   & $8.53$   & $9.21$   & $22.66$   & $55.34$    \\
    DECA \cite{Feng21} & $28.74$ & $29.24$ & $57.98$   & $38.17$  & $9.34$   & $11.66$   & $11.58$   & $16.69$   & $39.10$   & $84.06$   \\
    INORig \cite{Bai21}  & $27.68$ & $28.18$ & $55.86$   & $33.20$   & $23.25$   & $11.88$   & $\mathbf{4.86}$    & $9.75$    & $35.78$   & $80.54$    \\
    CEST \cite{Wen21} & $35.18$ & $12.14$ & $47.32$   & $29.92$  & $50.98$   & $38.77$   & $29.22$   & $23.62$   & $21.92$   & $46.57$     \\
    \hline
    Ours (BFM)      & $16.19$ & $15.33$ & $31.52$ & $21.82$ & $12.44$ & $\mathbf{6.48}$  & $5.69$   & $9.47$   & $16.67$  & $46.37$ \\
    Ours (AlbedoMM) & $17.72$ & $15.28$ & $33.00$ & $19.48$ & $15.50$  & $10.48$   & $8.42$   & $\mathbf{7.86}$  & $\mathbf{15.96}$ & $48.11$ \\
    Ours (BalancedAlb) & $\mathbf{13.87}$ & $\mathbf{2.79}$ & $\mathbf{16.67}$ & $\mathbf{18.41}$ & $11.90$   & $11.87$   & $11.20$   & $13.92$  & $16.15$ & $18.21$ \\
    \bottomrule
    \end{tabular}%
    } %
\end{table*}

\section{Unbiased Estimation via Scene Disambiguation Cues}

Resolving the ambiguity above requires additional information. 
While most methods in the literature work with a cropped face image, our key insight is that the larger image contains important disambiguation cues. %

\subsection{Model}

We begin by describing the image formation model employed throughout this work.

\paragraph{Geometry}
 We reconstruct geometry using the FLAME \cite{Li17} statistical model, which parameterizes a face/head mesh using identity $\shapecoeff \in \mathbb{R}^\shapedim$, pose $\posecoeff \in \mathbb{R}^{3\numjoints+3}$ (with $\numjoints=4$ the number of joints), and expression $\expcoeff \in \mathbb{R}^\expdim$ latent vectors.

\paragraph{Albedo} %
We use a low-dimensional linear model to represent diffuse albedo. 
To avoid the biases present in current publicly available models, we purchased 54 uniformly lit 3D head scans from 3DScanStore\footnote{\url{https://www.3dscanstore.com/}}, covering the full range of skin types as measured by the ITA score. 
We converted these into the $\uvsize\times\uvsize$ FLAME UV texture space, %
and used Principal Component Analysis (PCA) to learn a model that, given albedo parameters $\albedocoeffs \in \mathbb{R}^{\albedodim}$, outputs a UV albedo map $\albedo(\albedocoeffs) \in \mathbb{R}^{\uvsize \times \uvsize \times 3}$. The albedo image is reconstructed as $I_A = W(\albedo(\albedocoeffs)),$
where $W$ is a warping function that converts the UV map into camera space. 

\paragraph{Illumination}
We take the standard approach of approximating environment lighting using spherical harmonics (SH). 
Using this, 
the color of pixel $(i,j)$ is computed as
\begin{equation}
\label{eq:rendering}
    \begin{split}
    I_R(i,j) &= I_A(i,j) \cdot I_S(i,j)\\
     &= I_A(i,j) \cdot \sum_{k = 1}^{B^2}{\lighting_{k} \mathbf{H}_{k}(\normals(i,j))}\\
    &= I_A(i,j) \cdot \mathbf{\gamma} \cdot \mathbf{h}(i,j)\\
    \end{split}
\end{equation}
where $I_R(i,j)$ is the intensity at pixel $(i,j)$ (computed once for each channel); $B=3$ is the number of SH bands; $\mathbf{\gamma} =  ( \gamma_1, \dots, \gamma_{B^2} ) $ is the vector containing the SH parameters;   
$\mathbf{H}_{k},$ $k=1\ldots B^2,$ are the orthogonal SH basis functions, 
$\mathbf{h}(i,j) \in \mathbb{R}^{B^2}$ is the vectorised version of $\mathbf{H}_{k}(\normals(i,j))$, 
and $\normals$ is the normal image. 

Since the overall intensity of the image can be described by either the albedo $I_A$ or the shading $I_S$, we mitigate this fact by decomposing the SH coefficient vector of each color channel into a unit-scale vector and its norm:
\begin{equation}
\begin{split}
    I_R(i,j) &= || \mathbf{\gamma} || \cdot I_A(i,j) \cdot \frac{\gamma}{||\gamma||} \cdot \mathbf{h}(i,j).
\end{split}
\label{eq:sh_dec}
\end{equation}

The scale factor $ || \gamma ||$ can now be regarded as the overall light intensity (one value per RGB channel), while the unit-scale SH parameters $ \frac{\gamma}{||\gamma||}$ contain the light directional information. Given a fixed directional SH parameter, the scale is the key variable that modulates the skin tone. %
We  estimate this value  independent of the albedo and the normalized SH vector.

\subsection{TRUST Network}

An image of a face is typically only a small part of a larger image. 
Our key novelty here is to address the light/albedo ambiguity problem by leveraging \emph{scene cues} to regularize the ill-posed problem.
This is implemented by three design choices: (1) a novel \emph{scene consistency} loss, (2) a two-branch architecture that exploits the SH decomposition from Eq.~\ref{eq:sh_dec}, and (3) a light-conditioned albedo estimation network.

\paragraph{Scene consistency loss} First, we observe that a scene containing multiple faces can provide hints about the illumination. Taking inspiration from human color constancy, we can assume a single model of illumination for the entire scene, such that when we observe a variety of skin colors we know that the difference is due to albedo and not lighting\footnote{There are exceptions to this, such as a scenes where some faces are in shadow or where the lighting is high-frequency.}. 
 While other works have considered albedo consistency among different views as a cue for disambiguation \cite{Shi14, Tewari19}, \emph{light consistency} within an image has been mostly unexplored. 
We formalize the idea 
by requiring the SH parameters of different faces in a same  image to be close to each other. 
Specifically, given the set of all the estimated SH vectors $\{ \lighting^i \}_{i=\{1..N\}}$, where $N$ is the number of faces in the image, we penalize the difference between any two facial crops as $L_{sc} = ||P( \{ \lighting^i \} ) - \{ \lighting^i \} ||_1$, where $P(\cdot)$ is a random permutation function (illustrated in Fig.~\ref{fig:overview} right). 

However, we note that this loss alone cannot ensure a correct albedo reconstruction: it can only work when there is a variety of skin tones present in the scene, otherwise the estimated albedos can still be consistently brighter or darker than the true albedo.  

\paragraph{Two-branch architecture} As a complementary cue, we leverage the decomposition of SH into norm and direction described in Eq.~\ref{eq:sh_dec}, and exploit the fact that the global illuminant can provide additional information about the overall skin tone. To see why, consider the example of a dark-skinned person during daylight. The albedo can be correctly predicted as a dark skin under bright light, or incorrectly predicted as a light skin under dark light; the choice will depend largely on the bias of the prior. Yet, a coarse estimate of the global illuminant will reveal that the first case is the correct one.  
This provides a strong cue about where the output should lie in the skin color palette. 

We implement this observation by proposing a two-branch architecture, shown in Fig.~\ref{fig:overview}, left. 
First, the input image is passed through a \emph{scene light encoder} $E_{scene\_light}$ that predicts the norm of the scene spherical harmonics $ || \lighting ||$ for each RGB channel, thus capturing the overall light intensity and color. 
We next obtain facial crops using an off-the-shelf face detector (or ground-truth values at training time). 
From these crops, an \emph{albedo encoder} $E_{albedo}$ predicts the albedo parameters of the model $\albedocoeffs$, while a \emph{crop-based light encoder} $E_{face\_light}$ predicts the normalized SH vector $ \lighting' = \lighting /  || \lighting ||$. The outputs of $E_{scene\_light}$ and $E_{face\_light}$ are then combined to obtain the final SH prediction $\lighting$.

\paragraph{Light-conditioned albedo estimation} 
We note that during test time, the albedo estimator $E_{albedo}$ does not contain information about the scene, which can still introduce ambiguities. To address this, we propose to condition $E_{albedo}$ on the estimated light intensity. For this, we broadcast the intensity factor from the scene light encoder into an image of the same size as the facial crop, and concatenate it with the input as an additional channel.

\paragraph{Semi-supervised training} Given that unsupervised disentanglement cannot be solved without proper inductive biases \cite{Locatello18} or without a limited amount of potentially imprecise supervision \cite{Locatello19}, here we use a semi-supervised learning strategy. 
Specifically, we generate a synthetic training set of 50k images using 1170 scans acquired from Renderpeople\footnote{\url{https://renderpeople.com/}}, combined with 273 panoramic HDR images from Poly Haven\footnote{Note that these scenes are completely different from those used in the evaluation benchmark.}. We train the networks using a combined synthetic/real dataset to ensure generalization. 

\paragraph{Training} 
\modelname is trained
using the following loss function: 
\begin{equation}
    \mathcal{L} = \lambda_{pho} L_{pho} + \lambda_{sc} L_{sc} + \mathbb{I} \lambda_{SH} L_{SH} + \mathbb{I} \lambda_{alb} L_{alb}.
\end{equation}
Here, $L_{pho} = ||I - I_R||_1$ is the L1 photometric loss between the input image and rendered image, and $L_{sc}$ is the scene consistency loss. $\mathbb{I}$ is an indicator function with value 1 for supervised training data and 0 for real. 
When training with synthetic data we also employ an L1 loss $L_{SH} = ||\lighting - \tilde{\lighting}||_1^1$ between ground-truth SH coefficients $\lighting$ and the estimates  $\tilde{\lighting}$, 
as well as an L1 loss between the rendered albedo and ground-truth. 
Note that we include the self-supervised loss on synthetic data since the ground-truth light and albedo are only approximations to the physical ground-truth.
We set $\lambda_{pho}=10, \lambda_{sc}=10, \lambda_{SH} = 20, \lambda_{alb}=20$ to weight the loss terms based on validation-set performance.

To compute the photometric loss we need an estimate of the geometry and camera. For this we use a pre-trained state-of-the-art geometry estimation network, DECA~\cite{Feng21}, which provides FLAME shape and expression coefficients, as well as weak perspective camera parameters. This module is fixed during training.

\begin{figure*}[t]
    \offinterlineskip
    \centering
    \includegraphics[width=\textwidth]{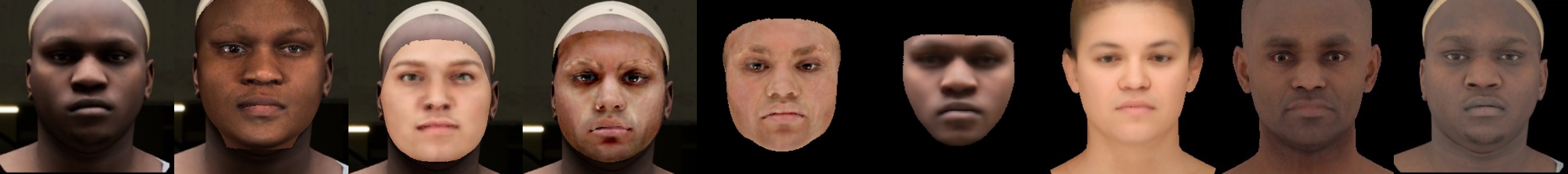}\\
    \includegraphics[width=\textwidth]{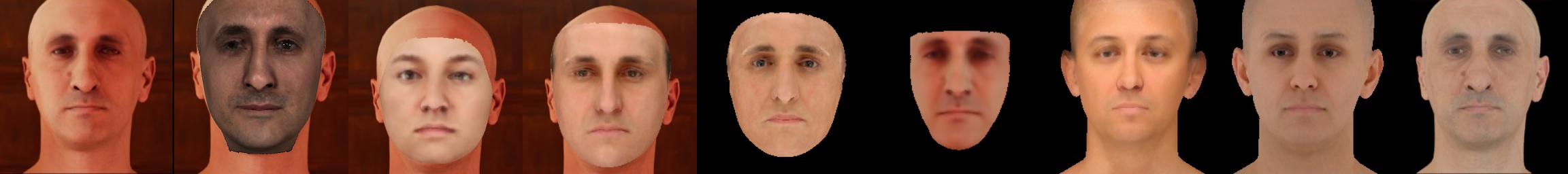}\\
    \includegraphics[width=\textwidth]{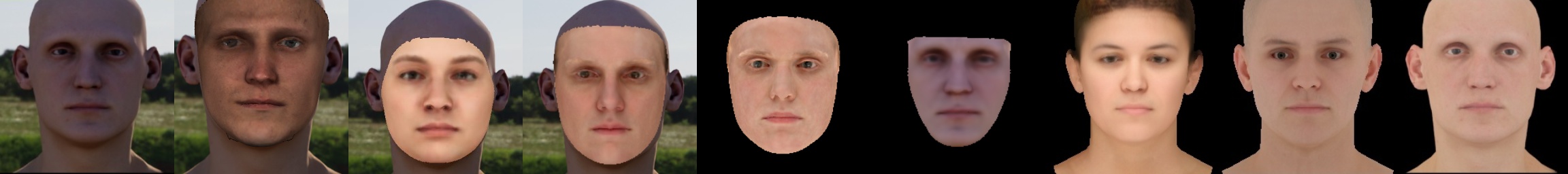}\\
    \includegraphics[width=\textwidth, trim={0 0.3cm 0 0}, clip=True]{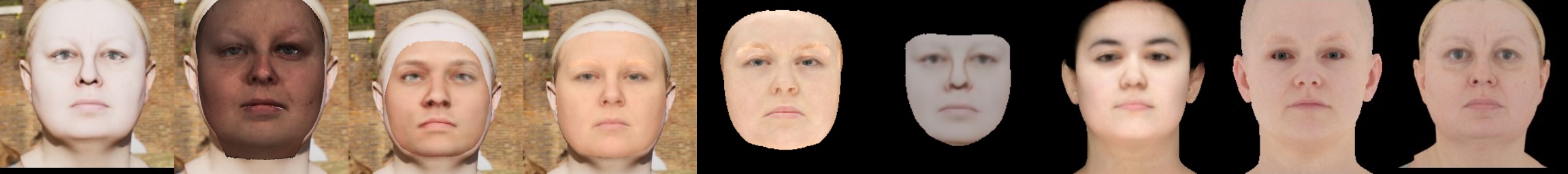}
    \includegraphics[width=\textwidth]{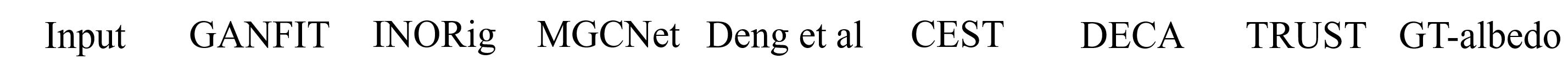}
	\caption{Comparison with recent face reconstruction methods on the proposed albedo benchmark. From left to right: input image, GANFIT~\cite{Gecer21}, INORig~\cite{Bai21}, MGCNet~\cite{Shang20}, Deng et al.~\cite{Deng19}, CEST~\cite{Wen21}, DECA~\cite{Feng21}, \modelname (ours) and ground-truth albedo rendering.
	}
    \label{fig:qual_bench}
\end{figure*}

\subsection{Implementation details}
We implemented our approach in PyTorch~\cite{Paszke2019PyTorchAI}, using the differentiable rasterizer from Pytorch3D~\cite{ravi2020pytorch3d} for rendering. We use Adam~\cite{Kingma14} as the optimizer with a learning rate of $1e-4$. All of the encoders use Resnet-50 architectures with input images of size $224 \times 224$ (in the case of full scene images, we first randomly crop a square and then resize). The UV space size is $\uvsize = 256$. To get shape, expression and camera parameters we use DECA~\cite{Feng21} to regress the FLAME parameters $\shapedim = 100$ and $\expdim = 50$. To train with real data, we use the subset of the OpenImages dataset~\cite{Krasin17} that contains faces. 

We apply a two-stage training strategy. The first stage employs fully supervised training with a batch size of 32 using the synthetic dataset for one epoch. The second stage uses semi-supervised training with a batch size of 48 and all proposed losses for 4 epochs. More details can be found in Sup.~Mat.
\begin{table}[t]
    \caption{
    {\bf Ablation study.} We show comparisons to the following alternatives: (1) light estimation from crops (``faceSH'') with self-supervised (self) and semi-supervised (semi) training sets; (2) SH intensity estimation from the scene, with SH directional estimation from crops (``fuseSH''); (3) fuseSH with scene consistency loss (``sc''); (4) fuseSH with conditioning (``cond''); (5) Ours: fuseSH with scene consistency and conditioning.
    } %
    \label{tab:ablations}
    \centering
    
    \begin{tabular}{@{}  lcccc@{} }
    \toprule
     Method & Avg. ITA $\downarrow$ & Bias $\downarrow$ & Score $\downarrow$ & MAE $\downarrow$ \\
    \midrule

    faceSH + self & $24.17$ &$11.46$ & $35.62$& $25.81$ \\
    faceSH + semi	& $14.70$ &$6.60$ & $21.31$& $17.64$\\
    \hline	
    fuseSH & $15.22$ & $11.08$ & $26.31$ & $\mathbf{17.02}$\\
    fuseSH + sc & $15.82$ & $3.50$ & $19.32$ & $20.37$\\
    fuseSH + cond & $\mathbf{14.16}$ & $6.56$ & $20.72$ & $17.05$\\
    Ours & $14.18$ & $\mathbf{2.63}$ & $\mathbf{16.81}$ & $19.08$\\
    \bottomrule
    \end{tabular}
\end{table}

\begin{figure}[htbp]
     \centering
     \begin{subfigure}{0.44\textwidth}
        \offinterlineskip
        \centering
        \includegraphics[width=\columnwidth]{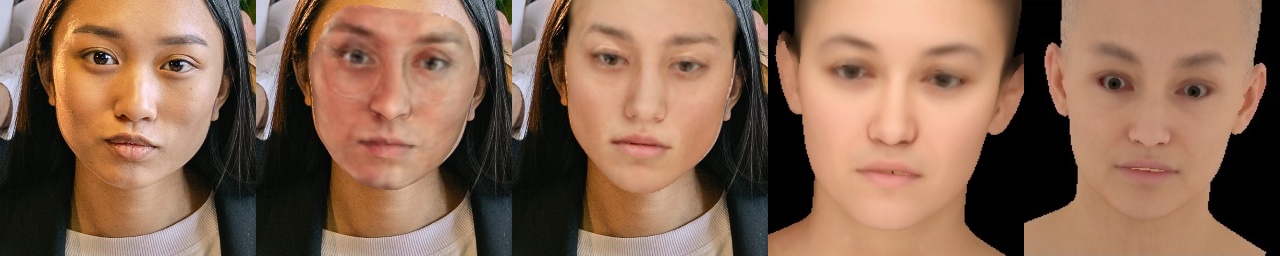}\\
        \includegraphics[width=\columnwidth]{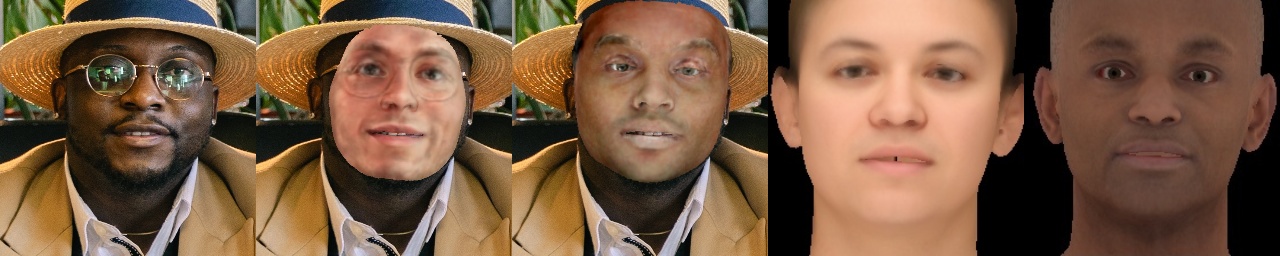}\\
        \includegraphics[width=\columnwidth]{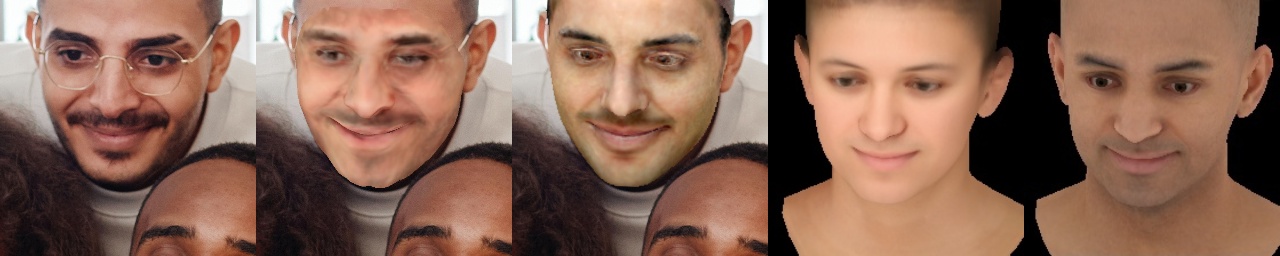}\\
        \includegraphics[width=\columnwidth]{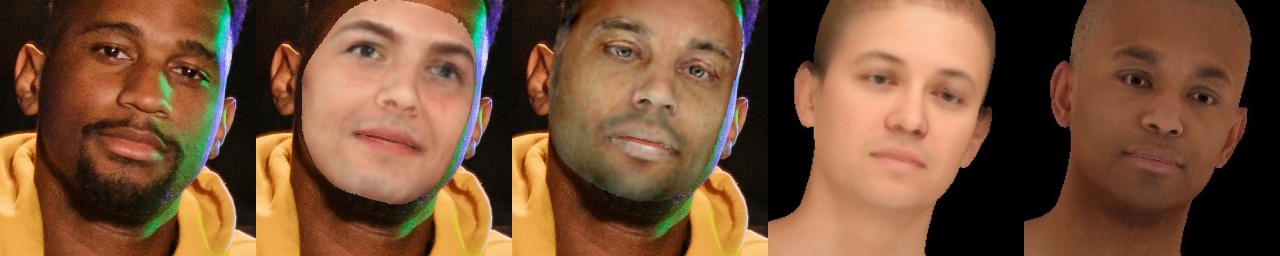}
        \caption{}
        \label{fig:qual_real}
     \end{subfigure}
     \begin{subfigure}{0.53\textwidth}
        \centering
        \includegraphics[width=\columnwidth]{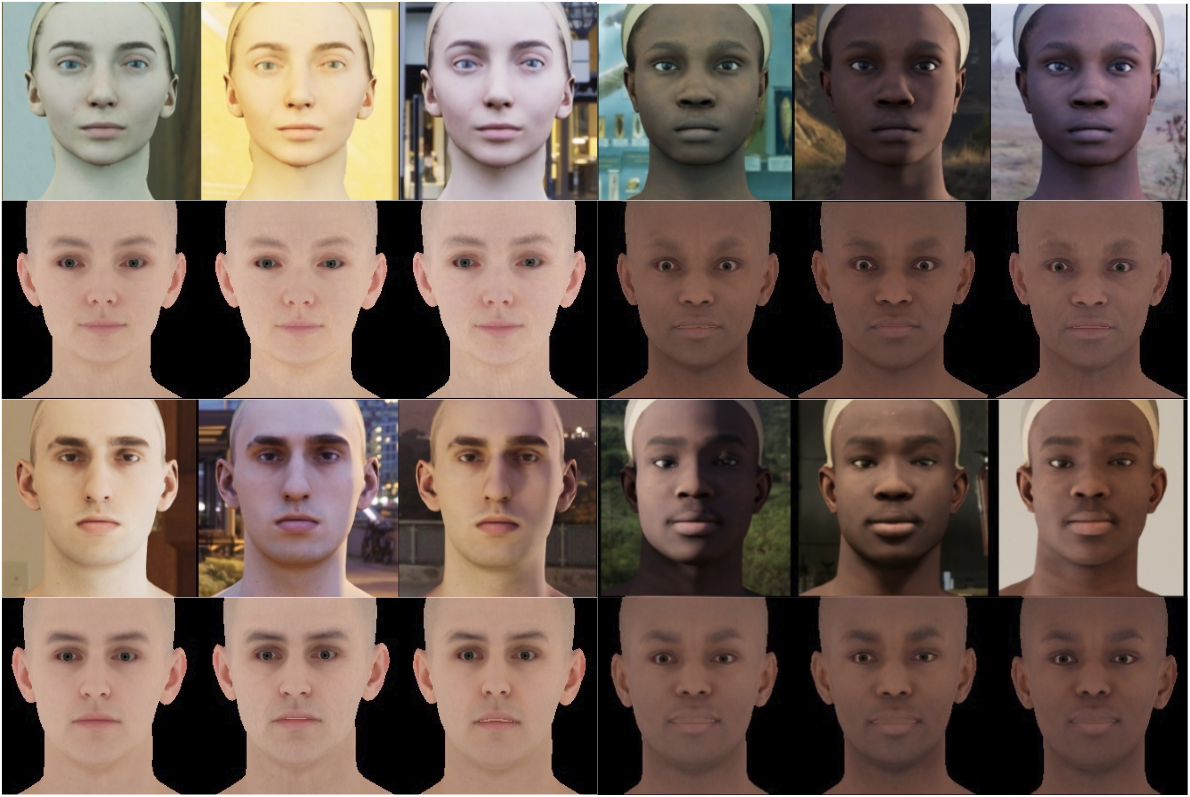}
        \caption{}
        \label{fig:consistent}
     \end{subfigure}
    \caption{
    \textbf{(a)} Qualitative comparisons on {\em real world images}. From left to right: 1) Input, 2) INORig~\cite{Bai21}, 3) MGCNet~\cite{Shang20}, 4) DECA~\cite{Feng21}, 5) \modelname (Ours). 
    \textbf{(b)} Given images of the same subject (from the \datasetname benchmark test set) under varying lighting (Rows 1 \& 3), \modelname outputs similar albedos (Rows 2 \& 4). 
    }
    \label{fig:three graphs}
\end{figure}

\section{Evaluation}
\label{sec:results}

We compare \modelname qualitatively and quantitatively with several SOTA methods.
MGCNet~\cite{Shang20}, Deng \etal~\cite{Deng19}, INORig~\cite{Bai21} and DECA~\cite{Feng21} use the Basel Face Model (BFM)  \cite{Paysan09} for albedo estimation; GANFIT~\cite{Gecer19} uses its own GAN-based appearance model; and CEST~\cite{Wen21} is a model-free approach.

\subsection{Qualitative results}
\label{sec:qual_eval}

We provide qualitative results on both real images (Fig.~\ref{fig:qual_real}) and synthetic images from the \datasetname benchmark (Fig.~\ref{fig:qual_bench}), showing faces with a variety of skin colors and scene illuminations. We observe that %
(1) GANFIT~\cite{Gecer19}, INORig~\cite{Bai21}, and DECA~\cite{Feng21} produce albedo maps %
with low variety, hence achieving low ITA values on specific skin types, but %
high ITA values %
for other skin types; (2) the model-free approach %
CEST~\cite{Wen21} bakes in a significant amount of light, since they cannot properly disentangle it from the albedo; (3) MGCNet~\cite{Shang20} and, to a certain degree, Deng \etal~\cite{Deng19} produce more diverse albedos, but since the BFM model does not include dark skin tones in the training dataset, these can only be encoded by extrapolating considerably, introducing noise. For more qualitative examples, see Sup.~Mat.

To evaluate robustness we apply \modelname on images of a same subject under different lighting, and with varying backgrounds. Fig.~\ref{fig:consistent} shows that the estimated albedo, per subject, is consistent across lighting and background variations, and that skin tones are well captured.

\subsection{Quantitative comparisons}
\label{sec:quant_eval}

We quantitatively compare several albedo estimation methods on the \datasetname benchmark as described in Sec.~\ref{sec:benchmark}. 
Table~\ref{tab:sota} shows that the proposed approach (using the balanced albedo model) outperforms existing  methods on all aggregated measures (i.e. Avg.~ITA, Bias, Score, and MAE), and produces a more uniform (low) error across skin types.
Note that while Deng \etal~\cite{Deng19}, INORig~\cite{Bai21}, and GANFIT~\cite{Gecer19} obtain the lowest scores for individual skin types, they have large errors for others, indicating their bias towards particular skin tones, which results in higher aggregated errors.

It is worth noting that, while shading information acts locally and provides gradients for geometry reconstruction, the skin tone is a global property of the albedo (low-frequency component). 
Hence, correct shape estimation does not necessarily imply good skin tone estimation (and vice-versa), which explains why methods such as DECA can achieve state-of-the-art results on shape estimation, even with strong regularization on the albedo coefficients that lead to incorrect skin tone estimates.

\subsection{Ablation studies}
\label{sec:ablations}

\noindent
\textbf{Effect of albedo model space.} 
We investigate how much a biased albedo space affects the final performance. To this end, we train two additional versions of our final network, but using instead the BFM albedo space \cite{Paysan09} and AlbedoMM \cite{Smith20}. 
Both model variants show large improvements on most skin types compared to prior work (see Tab.~\ref{tab:sota}). This demonstrates that addressing the light/albedo ambiguity is important for unbiased albedo estimation, and allows to push the albedo space to its representational limit.
Both model variants perform similarly (see Tab.~\ref{tab:sota}) in Avg.~ITA error and Bias scores (std), but around $20\%$ and $580\%$ higher than our final model with a balanced albedo space. Most errors of these two variants come from type V and VI, which shows the importance of using a balanced albedo model to cover the full range of skin tones.

\noindent
\textbf{Light and albedo estimation from facial crops (faceSH):} Here, both light and albedo encoders use facial crops as input, which is comparable to prior methods.
We evaluate this setting using self-supervised training and semi-supervised training, to test the importance of the synthetic vs in-the-wild training set for reducing racial bias. Results are shown in the first two rows of Table~\ref{tab:ablations}, where we see that all metrics are significantly improved with access to synthetic data.

\noindent
\textbf{Scene light estimation (fuseSH):} We next consider the case where the light intensity is estimated from the scene image, as in our approach, but with the following differences: (1) without conditioning and without scene consistency (fuseSH), (2) with scene consistency alone (fuseSH+sc), (3) with conditioning alone (fuseSH + cond). 
We observe in Table~\ref{tab:ablations} that each of these components improves a different aspect: while conditioning results in better skin color predictions, the scene consistency encourages fairer estimates. Our final model, which uses both scene-consistency and conditioning, achieves the lowest total score.

\noindent

\section{Conclusions}
This work presented initial steps towards unbiased estimation of facial albedo from images in the wild, using two main contributions. First, we built a new evaluation benchmark that is balanced in terms of skin type. We used this benchmark to highlight biases in current methods. We further proposed a new albedo estimation approach that addresses the light/albedo ambiguity problem, hence encouraging fairer estimates. Our solution is built on the idea that the scene image can be exploited as a cue to %
disambiguate light and albedo, resulting in more accurate predictions. The experimental results confirm that scene information helps to obtain fairer albedo reconstructions. 
We hope that this work will encourage the development of methods that are designed not just towards realism, but also towards fairness.

\bigskip
\paragraph{Acknowledgements}
We thank S. Sanyal for the helpful suggestions, O. Ben-Dov, R. Danecek, Y. Wen for helping with the baselines, N. Athanasiou, Y. Feng, Y. Xiu for proof-reading, and B. Pellkofer for the technical support.\\

\paragraph{Disclosure:} 
MJB has received research gift funds from Adobe, Intel, Nvidia, Meta/Facebook, and Amazon.  MJB has financial interests in Amazon, Datagen Technologies, and Meshcapade GmbH.  While MJB was a part-time employee of Amazon during a portion of this project, his research was performed solely at, and funded solely by, the Max Planck Society.
While TB is a part-time employee of Amazon, his research was performed solely at, and funded solely by, MPI. 

\clearpage
\bibliographystyle{splncs04}
\bibliography{bibliography}
\end{document}